\documentclass{article}
\usepackage{ijcai18}

\usepackage{hyperref}

\usepackage{times}
\usepackage{soul}
\usepackage{url}
\usepackage[utf8]{inputenc}
\usepackage[small]{caption}
\usepackage{color, colortbl}
\definecolor{mylightgray}{gray}{0.9}

\usepackage{fancyvrb}
\usepackage{multirow}
\usepackage{longtable}

\usepackage[colorinlistoftodos,prependcaption,textsize=tiny]{todonotes}

\usepackage{graphicx}
\usepackage{subcaption}
\usepackage{scrextend}

\title{Efficient Purely Convolutional Text Encoding}

\author{
  Szymon Malik\thanks{Equal contribution}, \ Adrian Lancucki\footnotemark[1], \ Jan Chorowski \\
  Institute of Computer Science \\
  University of Wrocław\\
  \texttt{
  szymon.w.malik@gmail.com,
  \{alan,jch\}@cs.uni.wroc.pl}
}

\begin{document}

\maketitle

\begin{abstract}
In this work, we focus on a lightweight convolutional architecture that creates fixed-size vector embeddings of sentences. 
Such representations are useful for building NLP systems, including conversational agents.
Our work derives from a recently proposed recursive convolutional 
architecture for auto-encoding text paragraphs at byte level.
We propose alternations that significantly reduce training time, the number of parameters, 
and improve auto-encoding accuracy. Finally, we evaluate the representations created by our model
on tasks from \textit{SentEval} benchmark suite, and show that it can serve as a better, yet fairly low-resource
alternative to popular bag-of-words embeddings.

\end{abstract}

\section{Introduction}
Modern conversational agents often make use of retrieval-based
response generation modules~\cite{Ram2018Conversational},
in which the response of the agent is retrieved from a curated database.
The retrieval can be implemented as similarity matching in a vector space,
in which natural language sentences are represented as fixed-size vectors.
Cosine and Euclidean distances typically serve as similarity measures.
Such approaches have been applied by participants of recent chatbot contests: The 2017 Alexa Prize~\cite{Pichl2018Alquist,Liu2017RubyStar,Serban2017Deep},
and The 2017 NIPS Conversational Intelligence Challenge~\cite{Chorowski2018ATalker,Yusupov2017Skill}.
Retrieval-based modules are fast and predictable.
Most importantly, they enable \textit{soft} matching between representations.
Apart from this straightforward application in dialogue systems,
sentence embeddings are applicable in downstream NLP tasks
relevant to dialogue systems.
Those include sentiment analysis~\cite{Pang2018Foundations},
question answering~\cite{Weissenborn2017FastQA}, censorship~\cite{Chorowski2018ATalker}, or intent detection.

Due to %
the temporal nature of natural languages, recurrent neural networks gained popularity
in NLP tasks.
Active research of different architectures led to great advances and eventually a shift
towards methods using the transformer architecture~\cite{Vaswani207Attention}
or convolutional layers~\cite{Bai2018AnEmpirical,Oord2017Parallel,Gehring2017Convolutional}
among researchers and practitioners alike. 
In this work, we focus on a lightweight convolutional architecture
that creates fixed-size representations of sentences. 

Convolutional neural networks have the inherent ability to detect local structures in the data.
In the context of conversational systems, their speed and memory efficiency
eases deployment on mobile devices, allowing fast response retrieval and better user experience.
We analyze and build on the recently proposed Byte-Level Recursive Convolutional
Auto-Encoder (BRCA) for text paragraphs~\cite{Zhang2018Byte-Level}, which is able to auto-encode
text paragraphs into fixed-size vectors, reading in bytes with
no additional pre-processing.

Based on our analysis, we are able to explain the behavior of this model,
point out possible enhancements, achieve auto-encoding accuracy improvements
and an order of magnitude training speed-up, while cutting down the number of parameters by over $70\%$.
We introduce a balanced padding scheme for input sequences and show that it
significantly improves convergence and capacity of the model.
As we find byte-level encoding unsuitable for embedding sentences,
we demonstrate its applicability in processing sentences at word-level.
We train the encoder %
with supervision on Stanford Natural Language Inference corpus~\cite{Bowman2015Large,Conneau2017Supervised} and investigate
its performance on various transfer tasks to assess quality of produced embeddings.

The paper is structured as follows:
in Section~\ref{sec:preliminaries} we introduce some of the notions that appear in the paper. 
Section~\ref{sec:related-work} discusses relevant work for sentence vector representations. Details of the 
architecture can be found in Section~\ref{sec:model-description}. Section~\ref{sec:model-analysis} 
presents the analysis of the auto-encoder and motivations for our improvements.
Section~\ref{sec:word-level-encoder} demonstrates supervised training of word-level sentence encoder,
and evaluates it on tasks relevant to conversational systems. Section~\ref{sec:conclusion} concludes the paper.

\section{Preliminaries}
\label{sec:preliminaries}
An information retrieval conversational agent selects a response from a fixed
set. Let $D = \{(i_k, r_k)\}_k$ be a set of conversational input-response pairs,
and $q$ be a current user's input.
Two simple ways of retrieving a response $r_k$ from available data are~\cite{Ritter2011Data,Chorowski2018ATalker}:
\vspace{-\topsep}
\begin{itemize}
  \setlength\itemsep{0em}
  \item return $r_k$ most similar to user's input $q$,
  \item return $r_k$ for which $i_k$ is most similar to $q$.
\end{itemize}
\vspace{-\topsep}
Utterances $i_k, r_k, q$ may be represented (embedded) as real-valued vectors. 

Many NLP systems represent words as points in a continuous vector space using word embedding 
methods~\cite{Mikolov2013Efficient,Pennington2014GloVe,Bojanowski2016Enriching}. They are calculated based on co-occurrence of words in large 
corpora. The same methods were applied to obtain sentence embeddings only with partial success, due to the 
combinatorial explosion of all possible word combinations, which make up a sentence. 
Instead, Recurrent Neural Networks (RNNs), autoregressive models that can process input sequences
of an arbitrary length, are thought to be a good method for handling variable-length textual data,
with Long Short-Term Memory network (LSTM)~\cite{Hochreiter1997Long} being the prime example of such.

Recently, RNNs have been reported to be successfully replaced 
by convolutional architectures~\cite{Bai2018AnEmpirical,Oord2017Parallel,Gehring2017Convolutional}. 
Convolutional neural networks are traditionally associated with computer vision
and image processing~\cite{Krizhevsky2012ImageNet,Redmon2015You}. They primarily consist of convolutional layers
that apply multiple convolutions to the input, followed by pooling layers
that are used for reducing the dimensionality
of the hidden state. 
Convolutional networks are efficient during training and inference:
they utilize few parameters and do not require sequential computations,
making hardware parallelism easy to use. Due to their popularity in image processing,
there are efficient implementations that scale well. 

Residual connection~\cite{He2015Deep} is a connection that adds an unchanged input
to the output of the layer or block of layers. During the forward pass it provides upper layers
with undistorted signal from the input and intermediate layers.
During the backward pass it mitigates vanishing
and exploding gradient problems~\cite{Hochreiter2001Gradient}. %

Batch Normalization~\cite{Ioffe2015Batch} (BN) applies normalization to all activations in every minibatch.
Typical operations used in neural networks are sensitive to changes in the range and magnitude of inputs.
During training the inputs to upper layers vary greatly due to changes in weights of the model.
Normalization of the signal in each layer has the potential to alleviate this problem.
Both BN and residual connections enable faster convergence
by helping with forward and backward flow of information.
They were also crucial in training our models.

\section{Related Work}
\label{sec:related-work}

There are many methods for creating sentence embeddings, the simplest being averaging
word-embedding vectors of a given sentence~\cite{Joulin2016Bag,Le2014Distributed}. 
SkipThought~\cite{Kiros2015Skip-Thought} generalizes idea of unsupervised learning of \textit{word2vec} word embeddings~\cite{Mikolov2013Efficient}.
It is implemented in the encoder-decoder setting using LSTM networks.
Given a triplet of consecutive sentences $(s_{i-1}, s_i, s_{i+1})$, the encoder creates a fixed-size 
embedding vector of the sentence $s_i$, and the decoder tries to generate sentences
$s_{i-1}$ and $s_{i+1}$ from this representation. 
SkipThought vectors have been shown to preserve syntactic and semantic properties
~\cite{Kiros2015Skip-Thought}, so that their similarity is represented in the embedding space. 

InferSent model~\cite{Conneau2017Supervised} shows that training embedding systems with supervision on a
natural language inference task may be superior to an unsupervised training. 
Recently, better results were obtained by combining supervised learning on an auxiliary
natural language inference corpus with learning to predict the correct response on a conversational data 
corpus~\cite{Yang2018Learning}.

\section{Model Description}
\label{sec:model-description}

\begin{figure}[tb]
\begin{subfigure}[b]{0.49\columnwidth}
  \begin{center}
    \includegraphics[scale=0.54]{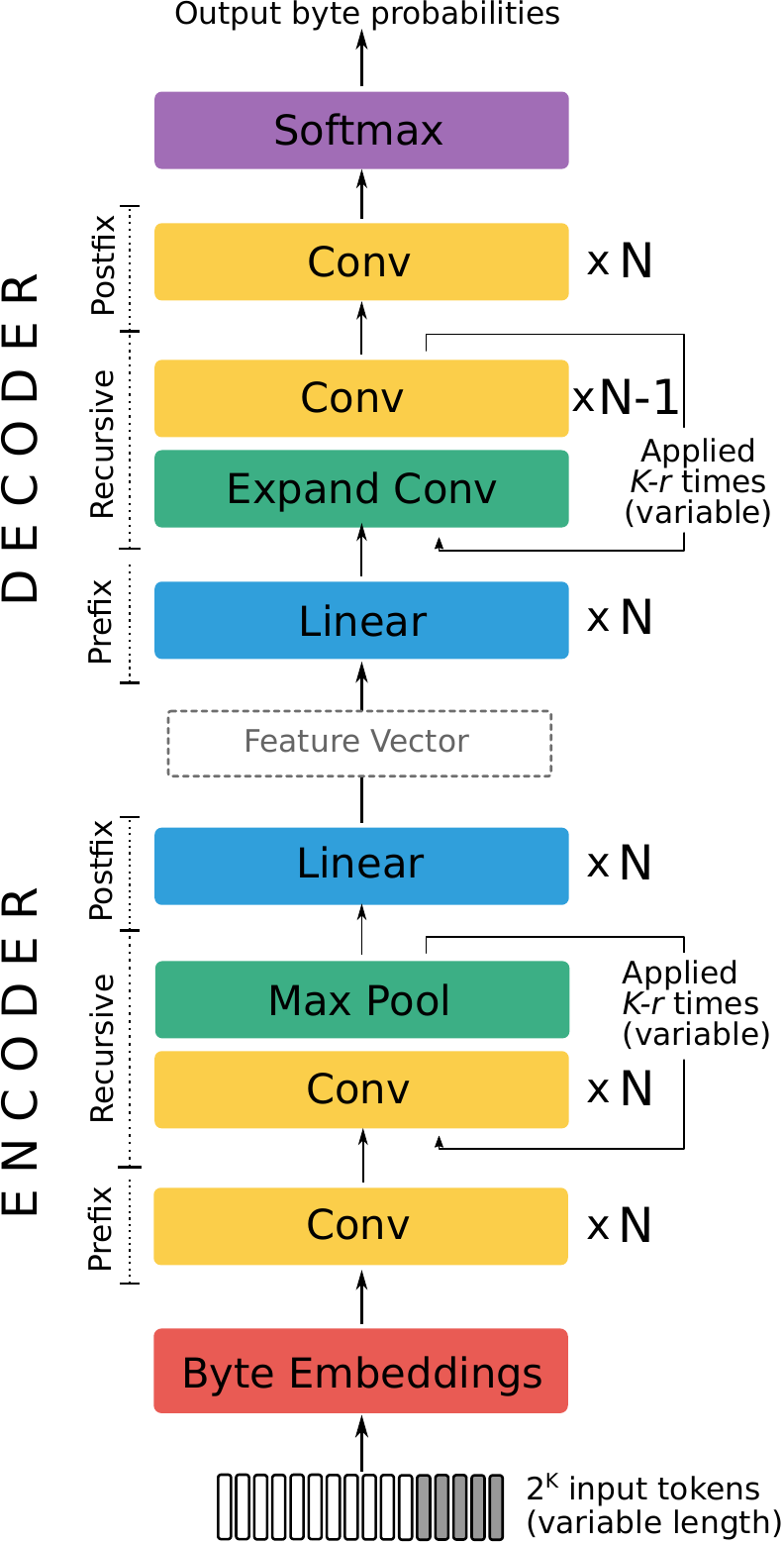}
     \caption{BRCA}
  \end{center}
\end{subfigure}
\begin{subfigure}[b]{0.49\columnwidth}
  \begin{center}
    \includegraphics[scale=0.54]{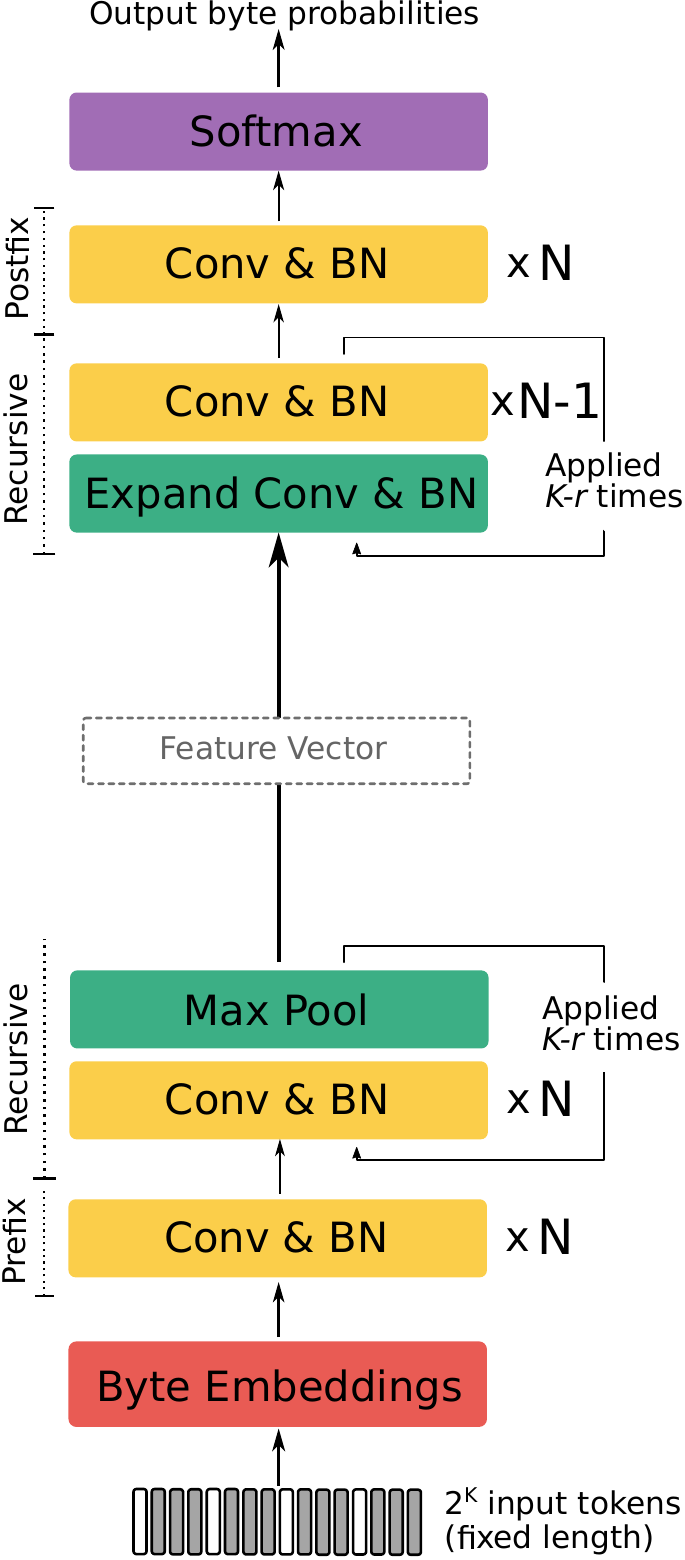}
    \hspace{-0.6cm}
    \caption{Our model}
  \end{center}
\end{subfigure}
  \caption{Structural comparison of Byte-Level
    Recursive Convolutional Auto-Encoder
    (BRCA) and our model. Dark boxes indicate input padding. BRCA pads the input form the right to the nearest power
    of two. We pad the input evenly to a fixed-size vector. Our model does not have
    postfix/prefix groups with linear layers and uses Batch Normalization (BN) after every layer.}
  \label{fig:models}
\end{figure}

Our model builds on the Byte-Level Recursive Convolutional Auto-Encoder~\cite{Zhang2018Byte-Level}
(BRCA). Both models use a symmetrical encoder and a decoder.
They encode a variable-length input sequence as a fixed-size latent representation,
by applying a variable number of convolve-and-pool stages
(Figure~\ref{fig:models}).
Unlike in autoregressive models, the recursion is not applied
over the input sequence length, but over the depth of the model.
As a result, each recursive step processes all sequence elements in parallel.

\subsection{Encoder}

Our encoder uses two operation groups from the BRCA: a prefix group and a recursive group.
The \textbf{prefix group} consists of $N$ temporal convolutional layers,
and the \textbf{recursive group} consists of $N$ temporal convolutional layers
followed by a max-pool layer with kernel size $2$.
For each sequence, the prefix group is applied only once,
while the recursive group is applied multiple times, sharing weights between all applications.
All convolutional layers have $d=256$ channels, kernels of size $3$ and are organized into residual blocks,
with 2 layers per block, ReLU activations, residual connections, and Batch Normalization
(see Section~\ref{sec:Ioffe2015Batch} for details).

The encoder of our model reads in text,
by sentence or by paragraph, as a~sequence of discrete tokens (e.g. bytes, characters, words).
Each input token is embedded as a fixed-length $d$-dimensional vector.
Unlike~\cite{Zhang2018Byte-Level}, where input sequence is zero-padded to the nearest power of~2,
we pad the input to a fixed length $2^K$,
distributing the characters evenly
across the input. We motivate this decision by the finding
that the model zero-padded to the nearest power of~2
does not generalize to sentences longer
than those seen in the training data (see Section~\ref{sec:capacity}).

First, the prefix group is applied, which retains the dimensionality
and the number of channels, i.e., sequence length and embedding size.
Let $d \cdot 2^r$ be the dimensionality of the latent code output by the encoder.
The encoder then applies the recursive group $K-r$ times. Note that with $r$ one may control 
size of a latent vector (the level of the compression).
With the max-pooling, every application halves
the length of the sequence. Weights are shared between applications.
Finally, the encoder outputs a latent code of size $d \cdot 2^r$.
Unlike~\cite{Zhang2018Byte-Level}, we do not apply any linear layers after recursions. 
Our experiments have shown that they slightly degrade the performance of the model,
and constitute the majority of its parameters.

\subsection{Decoder}

The decoder acts in reverse. First, it applies a 
\textbf{recursive group} consisting of a convolutional layer which doubles the number of channels
to $2d$, which is followed by an expand transformation~\cite{Zhang2018Byte-Level} and $N-1$ convolutional layers.
Then it applies a \textbf{postfix group} of $N$ temporal convolutional layers.
Similarly to the encoder, the layers are organized into residual blocks with ReLU
activations, Batch Normalization, and have the same dimensionality and kernel size.
We double the size of input in the residual connection, which bypasses the first two convolutions
of the recursive group, by stacking it with itself.
We found it crucial for convergence speed to use only residual blocks in the network,
also in the expand block.

The decoder applies its recursive group $K-r$ times. Each application doubles the numbers of channels,
while the expand transformation reorders and reshapes the data
to effectively double the length of the sequence while the number of channels is unchanged and equals $d$.
The postfix group processes a tensor of size $2^K\times d$ and retains its dimensionality.
The output is interpreted as $2^K$ probability distributions over possible output elements.
Adding an output embedding layer, either tied with input embedding layer or separate,
slowed down training and did not improve the results.
At the end, a Softmax layer is used to compute output probabilities over possible bytes.
Note that output probabilities are independent from one another conditioned on the input.

\section{Model Analysis}
\label{sec:model-analysis}

In this section we justify our design choices through a series of experiments
with the BRCA model and report their outcomes.\footnote{Source code of our models is available: \\ \url{https://github.com/smalik169/recursive-convolutional-autoencoder}}

\subsection{Data}

In order to produce comparable results,
we prepared an English Wikipedia dataset with similar sentence length distribution to~\cite{Zhang2018Byte-Level}.
Namely, we took at random 11 million sentences from an English Wikipedia dump extracted with
WikiExtractor\footnote{\url{https://github.com/attardi/wikiextractor}},
so that their length distribution would roughly match that of~\cite{Zhang2018Byte-Level} (Table~\ref{tab:wiki}).
In experiments with random data, we generate random strings of \texttt{a-zA-Z0-9} ASCII
characters.

\begin{table}[h]
\caption{Lengths of paragraphs in the English Wikipedia dataset}
\label{tab:wiki}
\begin{center}
\begin{tabular}{|c|c|}
  \hline
  Length & Percentage\\
  \hline
  \hline
  4-63 B     & 35\% \\
  \hline
  64-127 B   & 14\% \\
  \hline
  128-255 B  & 20\% \\
  \hline
  256-511 B  & 18\% \\
  \hline
  512-1023 B & 14\% \\
  \hline
\end{tabular}
\end{center}
\end{table}

\subsection{Model Capacity}
\label{sec:capacity}

Natural language texts are highly compressible due to their low entropy,
which results from redundancy of the language~\cite{Levitin1994Entropy}.
In spite of this, the considered models struggle to
auto-encode 1024-byte short paragraphs into 1024-float latent vectors, which are
4096-byte given their sheer information content.
Transition from discrete to continuous representation and inherent inefficiency
of the model are likely to account for some of this overhead.

One can imagine an initialization of weights that, given the over-capacitated
latent representation, would make the network perform identity for
paragraphs up to 128 bytes long\footnote{When max-pooling is replaced by convolution
with stride 2 and kernel size 2}.
We confirmed those speculations experimentally, training models on paragraphs
of random printable ASCII characters,
namely random strings of \texttt{a-zA-Z0-9} symbols (Table~\ref{tab:identity}).
\begin{table}[t]
\caption{Learning identity by training on random sequences of ASCII characters
  of different length. Accuracy is presented for BRCA (N=8) model.}
\label{tab:identity}
\begin{center}
\begin{tabular}{|c|c|c|}
    \hline
	Training Lengths & Test Length & Accuracy \\
	\hline\hline
    4 -- 128                  & 128 & $99.81\%$ \\
    \hline
    \multirow{3}{*}{4 -- 512} & 128 & $60.79\%$ \\
    \cline{2-3}
						      & 256 & $22.99\%$ \\
    \cline{2-3}
						      & 512 & $9.81\%$ \\
	\hline
\end{tabular}
\end{center}
\end{table}
The empirical capacity of our model is 128 bytes, which sheds light
on the amount of overhead. This model has to be trained on paragraphs longer than 512 
bytes in order to learn useful, compressing behavior given a 1024-float latent representation.

\subsection{Generalization to Longer Sequences}
Auto-encoding RNN models such as LSTM are known to deteriorate
gradually with longer sequences~\cite{Kyunghyun2014Properties,Chorowski2015Attention}.
We trained a BRCA model ($N=2$) and a LSTM encoder-decoder
network with hidden size $256$. Both models were trained
on sentences of length up to $128$ bytes and evaluated on unseen data.
The LSTM model did not perfectly learn the identity function,
even though it was solving an easier task of predicting the character given the correct prefix. 
However, the LSTM model generalized much better on longer sequences,
where performance of BRCA deteriorated rapidly (Table~\ref{tab:generalization}).
\begin{table}[t]
\caption{Comparison of the ability of BRCA and LSTM encoder-decoder
  to learn an identity function and generalize to unseen data.
  Values represent byte-level decoding accuracy.
  Note that the LSTM decoder has the advantage
  of always being primed with the correct prefix sequence.}
\label{tab:generalization}
\begin{center}
\begin{tabular}{|c|c|c|}
  \hline
    Lengths (bytes) & BRCA (N=2) & LSTM-LSTM \\
  \hline\hline
    9-16 & 97.06\% & 91.17\% \\
  \hline
    17-32 & 97.96\% & 90.20\% \\
  \hline
    33-64 & 97.45\% & 91.72\% \\
  \hline
    65-128 & 83.56\% & 86.34\% \\
  \hline
    129-256 & 11.66\% & 72.88\% \\
  \hline
    257-512 & 8.05\% & 58.80\% \\
  \hline
\end{tabular}
\end{center}
\end{table}

\subsection{Balanced Padding of Input Sequences}

We found BRCA difficult to train. The default hyperparameters given by the authors~\cite{Zhang2018Byte-Level}
are single-sample batches, SGD with momentum $0.9$, and a small learning rate
$0.01$ with 100 epochs of training.
In our preliminary experiments,
increasing the batch size by batching paragraphs of the same length improved convergence
on datasets with short sentences (mostly up to 256 bytes long),
but otherwise deteriorated on the Wikipedia dataset, where roughly 50\% paragraphs
are longer than 256 bytes.
We suspect that the difficulty lies in the difference of the underlying tasks:
long paragraphs require compressive behavior,
while short ones merely require learning the identity function.
Updating network parameters towards
one tasks hinders the performance on the others,
hence the necessity for careful training.

In order to blend in both tasks, we opted for padding input sequences
into fixed-length vectors. We find it sensible to fix maximum length of input sentence,
since the model does not generalize to unseen lengths anyway.
Variable length of input in BRCA does save computations, 
however we found fixing input size to greatly improve training time,
despite the overhead.

\begin{figure}[t]
\begin{subfigure}[b]{0.45\columnwidth}
  \begin{center}
    \includegraphics[width=0.5\textwidth]{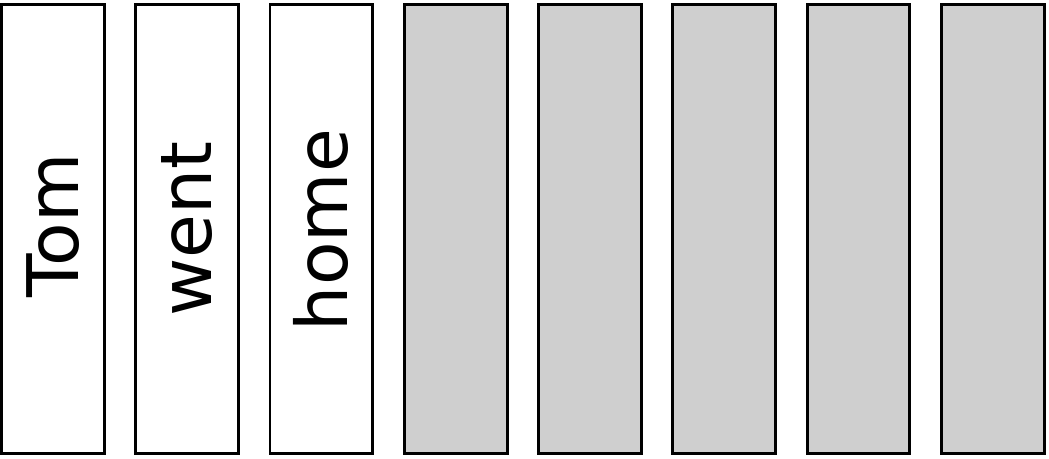}
     \caption{Unbalanced padding to sequence of length 8}
  \end{center}
\end{subfigure}
\hspace{0.2cm}
\begin{subfigure}[b]{0.45\columnwidth}
  \includegraphics[width=\textwidth]{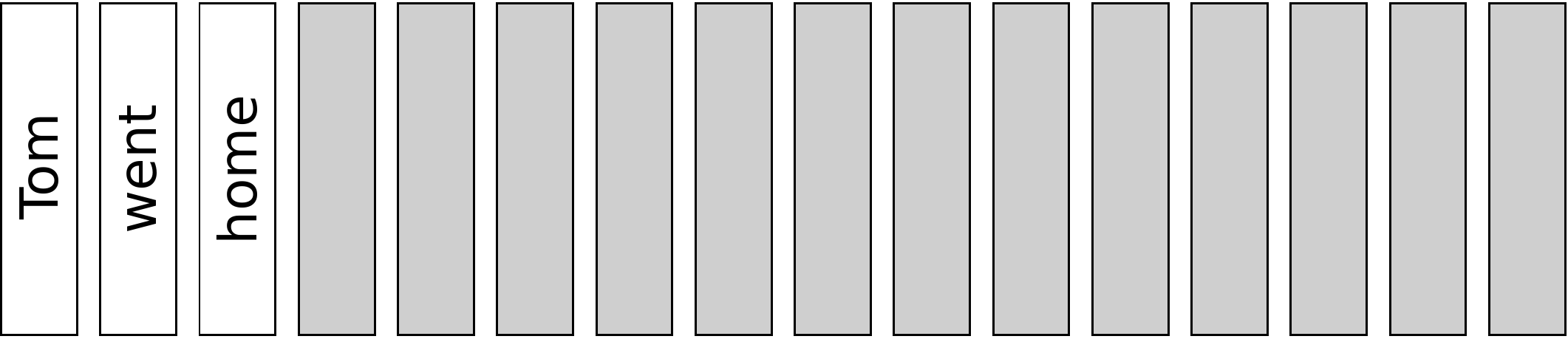}
  \caption{Unbalanced padding to sequence length $16$}
\end{subfigure}
\\
\begin{subfigure}[b]{0.45\columnwidth}
  \begin{center}
    \includegraphics[width=0.5\textwidth]{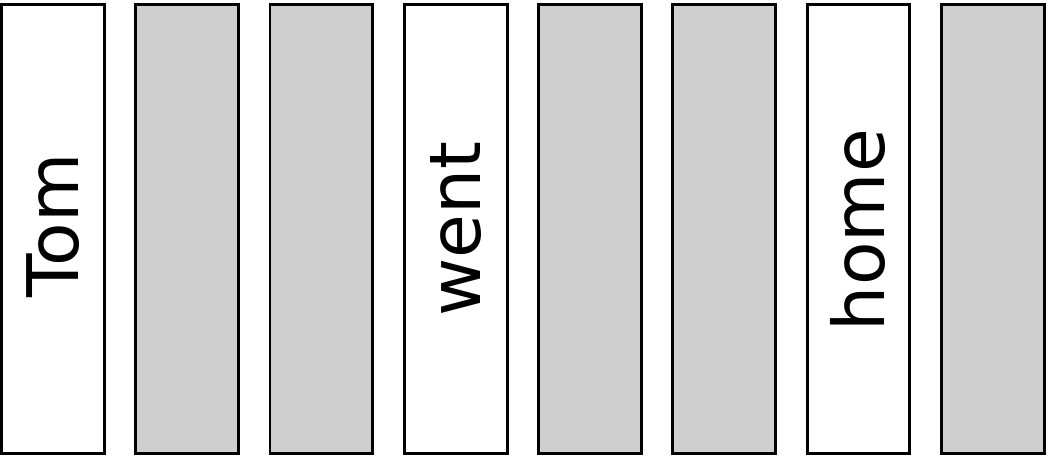}
    \caption{Balanced padding to sequence of length 8}
  \end{center}
\end{subfigure}
\hspace{0.2cm}
\begin{subfigure}[b]{0.45\columnwidth}
  \includegraphics[width=\textwidth]{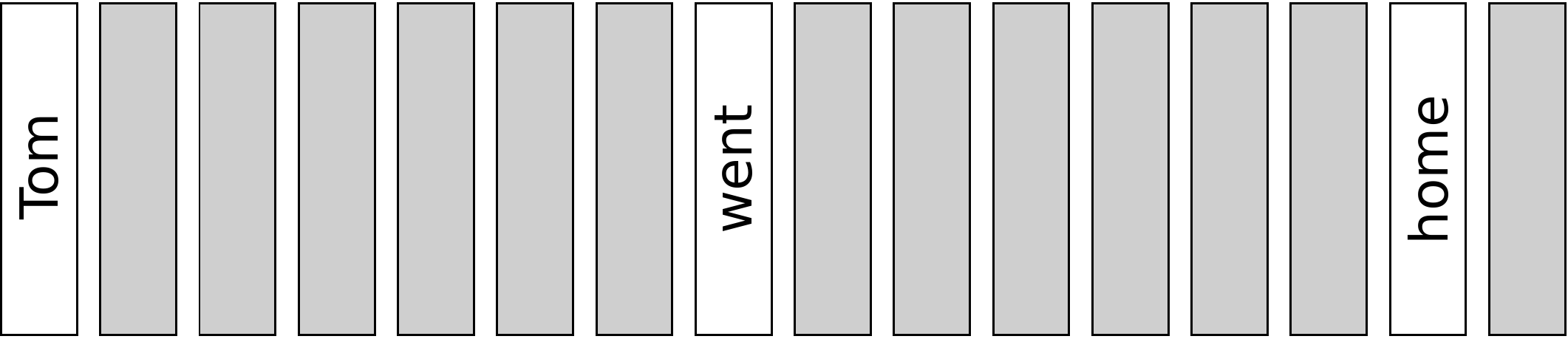}
  \caption{Balanced padding to sequence length $16$}
\end{subfigure}
  \caption{Unbalanced and balanced padding of an input sequence
    to a fixed-length sequence. Grey boxes are (zero) padding,
    white boxes are input embedding vectors.}
  \label{fig:padding}
\end{figure}

In order to make the tasks more similar, we propose balanced padding
of the inputs~(Figure~\ref{fig:padding}). Instead of padding from the right
up to $2^K$ bytes, we pad to the nearest power of~2 and
distribute the remaining padding equally in between the bytes.
We hypothesized that it could free convolutional layers from the burden of
propagating the signal from left to right in order to fill the whole latent vector,
as it would be the case, e.g., when processing a 64-byte paragraph padded
with 960 empty tokens from the right to form 1024-byte input.
Empirically, this trades additional computations for better convergence characteristics.

\begin{figure*}[t]
  \begin{subfigure}[b]{1.0\linewidth}
    \begin{center}
    \includegraphics[width=0.85\linewidth]{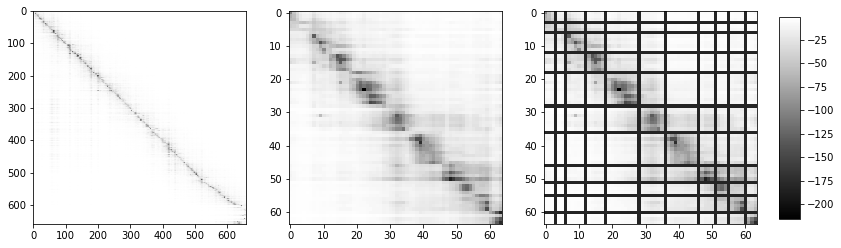}\\
    \end{center}
    \caption{Input sentence: \textit{One of Lem's major recurring themes, 
             beginning from his very first novel, "The Man from Mars" (\ldots)}}
  \end{subfigure}
  \begin{subfigure}[b]{1.0\linewidth}
    \begin{center}
    \includegraphics[width=0.85\linewidth]{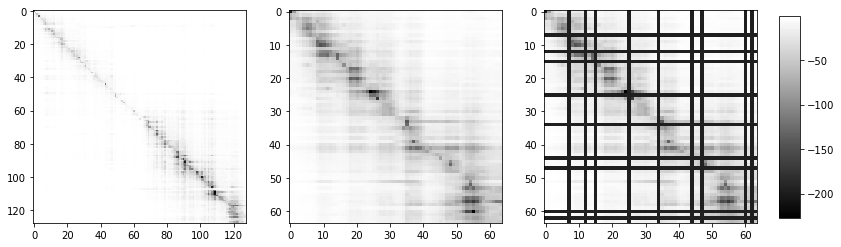}
    \end{center}
    \caption{Input sentence: \textit{Typical fuel is denatured alcohol,
             methanol, or isopropanol (\ldots)}}
  \end{subfigure}
  \caption{Input-output byte relations (X axis vs. Y axis)
           as indicated by the method of Integrated
           Gradients~\protect\cite{Sundararajan2017Axiomatic}
           with 50 integration points.
           The plots correspond to (a) 659-byte, and (b) 128-byte
           Wikipedia paragraphs.
           The leftmost plots show relations between all input-output bytes,
           the middle plots for the first 64 bytes.
           The rightmost plots also plot spaces.
           Dark shades indicate strong relations - those lay along diagonals
           and do not cross word and phrases boundaries.}
  \label{fig:heatmaps}
\end{figure*}

\subsection{Batch Normalization}
\label{sec:Ioffe2015Batch}

Fixed-length, balance-padded inputs allow easy mixing of paragraphs of different
lengths within a batch, in consequence allowing raising the batch size, 
applying Batch Normalization and raising the learning rate.
This enables a significant speed-up in convergence and better
auto-encoding accuracy (see Section~\ref{sec:accuracy}).
However, the statistics collected by BN layers differ
during each of the $K-r$ recursive steps,
even though the weights of convolutions in the recursive
layers are shared. 
This breaks auto-encoding during inference time, when BN layers
have fixed mean and standard deviation collected over a large dataset.
We propose to alleviate this issue by either: a) 
collecting separate statistics for each recursive application
and each input length separately,
or b) placing a paragraph inside a batch of data drawn from the training corpus
during inference and calculating the mean and the standard deviation on this batch.
We also experimented with the instance normalization~\cite{Ulyanov2016Instance},
which performs the normalization of features of a single instances,
rather than of a whole minibatch.
We have found that the instance normalization improved greatly upon the baseline model
with no normalization, but performed worse than batch normalization.

BRCA has been introduced with linear layers in the postfix/prefix groups of the encoder/decoder.
In our experiments, removing those layers from the vanilla BRCA
lowered accuracy by a few percentage points.
Conversely, our model benefits from not having
linear layers. We observed faster convergence and better accuracy without them,
while reducing the number of parameters from $23.4$ million to $6.67$ million.

\subsection{Auto-Encoding Performance}
\label{sec:accuracy}

Our training setup is comparable with that of BRCA~\cite{Zhang2018Byte-Level}.
In each epoch, we randomly select $1$ million sentences from the training
corpus. We trained using SGD with momentum $0.5$ in
batches of $32$ paragraphs of random length, balanced padded to $2^r=1024$
tokens, including a special end-of-sequence token.
The training was run for $16$ epochs, and learning rate was multiplied by $0.1$
every epoch after the $10$th epoch.
The model suffered from the exploding gradient problem~\cite{Hochreiter2001Gradient},
and gradient clipping stabilized the training, enabling even higher learning rates.
With clipping, we were able to set the learning rate as high as $30.0$,
cutting down training time to as low as 5 epochs.

\begin{figure}[t]
  \includegraphics[width=1.03\linewidth]{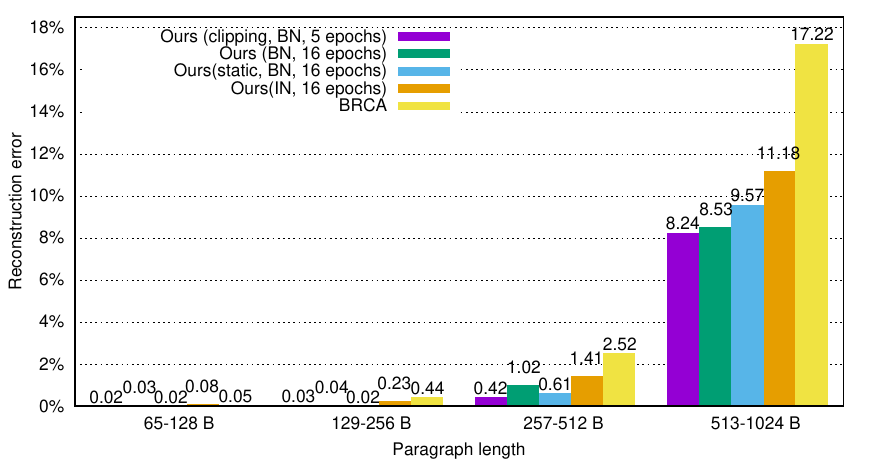}
  \caption{Decoding errors on unseen data for our best models
      ($N=8$, no linear layers)
      with balanced input padding to a sequence of size $1024$
      compared with Byte-Level Recursive Convolutional Auto-Encoder (BRCA)}
  \label{fig:err_by_len}
\end{figure}

Figure~\ref{fig:err_by_len} shows auto-encoding
error rate on the test set by sentence length.
Our best model achieved $1.5\%$
test error, computed as average byte-level error on the English Wikipedia dataset.

Finally, we were able to train a static version of our model
(i.e., with no shared weights in the recursion group) in comparable time,
closing a huge gap in convergence of recursive and static models in vanilla BRCA.

\begin{figure*}[htb]
\begin{center}
\begin{subfigure}[b]{0.9\textwidth}
{\scriptsize
\begin{center}
\begin{BVerbatim}[commandchars=+\{\}]

In :  9H3cxn4RIRUnOuPymw28dxUoA060LQ3heq1diKcbiUoinkzDjxucnE3Hk7FEFwHjzcTlOrhPUp3kgt9y8VAaw1sYpjPO9N5Cv4IAn
Out:  9H3cxn4RIRUnOuPymw28dxUoA060LQ3heq+textcolor{red}{+textbf{d}}diKcbiUoinkzDjxucnE3Hk7FEFwHjzcTlOrhPUp3kgt9y8VAaw1sYpjPO9N5Cv4IAn

\end{BVerbatim}
\end{center}
}
\caption{Random string of characters (under 128 bytes)}
\end{subfigure}
\end{center}

\begin{center}
\begin{subfigure}[b]{0.9\textwidth}
{\scriptsize
\begin{center}
\begin{BVerbatim}[commandchars=+\{\}]

In :  Lorsque ce m\xc3\xa9lange de cultures mondiales doit donner une signification au fa
      it d'\xc3\xaatre humain, ils r\xc3\xa9pondent avec 10,000 voix diff\xc3\xa9rentes.
Out:  Lorsque ce m\xc3\xa9lange de cultures mondiales doit donner une signification au fa
      it d'\xc3\xaatre humain, ils r\xc3\xa9pondent avec 10,000 voix diff\xc3\xa9rentes.


In :  When we first sequenced this genome in 1995, the standard of accuracy was one error per 10,000 base pairs.
Out:  When we first sequenced this genome in 1995, the standard of accuracy was one error per 10,000 base pairs.
\end{BVerbatim}
\end{center}
}
\caption{French and English sentences (under 256 and 128 bytes respectively)
}
\end{subfigure}
\end{center}

\begin{center}
\begin{subfigure}[b]{0.9\textwidth}
{\scriptsize
\begin{center}
\begin{BVerbatim}[commandchars=+\{\}]

In :  Lorsque ce m\xc3\xa9lange de cultures mondiales doit donner une signification au fa
      it d'\xc3\xaatre humain, ils r\xc3\xa9pondent avec 10,000 voix diff\xc3\xa9rentes.L
Out:  L+textcolor{red}{+textbf{a}}r+textcolor{red}{+textbf{kere}} +textcolor{red}{+textbf{d}}e +textcolor{red}{+textbf{Bu}}\xa9   lan+textcolor{red}{+textbf{d}}e +textcolor{red}{+textbf{by mor}}tures mondiales +textcolor{red}{+textbf{le}}i+textcolor{red}{+textbf{d}} d+textcolor{red}{+textbf{u}}nner +textcolor{red}{+textbf{o}}ne +textcolor{red}{+textbf{m}}ignification +textcolor{red}{+textbf{or Z}}a
      +textcolor{red}{+textbf{an '}}'\xc3+textcolor{red}{+textbf{u   a}}r+textcolor{red}{+textbf{i R}}umain+textcolor{red}{+textbf{s and pe   e   racha}}nt +textcolor{red}{+textbf{op}}e+textcolor{red}{+textbf{n (}}0,000 +textcolor{red}{+textbf{se}}i+textcolor{red}{+textbf{d}} di+textcolor{red}{+textbf{sar   s   s}}en+textcolor{red}{+textbf{g}}e+textcolor{red}{+textbf{. M}}

In :  When we first sequenced this genome in 1995, the standard of accuracy was one error 
      per 10,000 base pairs.When we first sequence d this genome in 1995, the standard of
Out:  When we first sequenced this genome in 1995, the standard of accuracy was one error
      per 10,000 base pair+textcolor{red}{+textbf{.}}.+textcolor{red}{+textbf{PW}}e+textcolor{red}{+textbf{w}} we first sequence d this genome in 1995, the standard of                                       
\end{BVerbatim}
\end{center}
}
\caption{French and English sentences concatenated 4 times to form a longer input (only a prefix is shown above)
}
\end{subfigure}
\end{center}                                        
  \caption{
      Auto-encoding capabilities of the model with errors marked in \textcolor{red}{\textbf{bold red}}.
      The model was trained only on English Wikipedia paragraphs.
      On short sequences, our model performs close to an identity function.
      On longer ones, it seems to correctly auto-encode only
      English paragraphs. Note that the model tries to map French words into English
      ones (\texttt{avec} $\rightarrow$ \texttt{open}, \texttt{une} $\rightarrow$ \texttt{one}).
      We observed a similar behavior on other languages as well.
  }
  \label{fig:samples}
\end{figure*}

\begin{table*}[t]
\caption{
Results for word-level sentence encoders.
We compare bag-of-words (BoW), i.e. averaged word embeddings, WRCE - the encoder from Zhang and LeCun's model on word-level, our word-level model with balanced padding to 64 elements (Ours), and an ensemble of our model and BoW (Ours + BoW) for various supervised (classification accuracy) and unsupervised (Pearson/Spearman correlation coefficients) tasks.}
\label{table:word-level}
\begin{center}
\begin{tabular}{l|c|c|c|c|}
    \cline{2-5}
    & \multicolumn{4}{c|}{Model} \\
    \hline
	\multicolumn{1}{|l|}{Task (dev/test acc\%)} & BoW & WRCE & Ours & Ours + BoW \\
	\hline\hline
    \multicolumn{1}{|l|}{SNLI} & 67.7 / 67.5 & 82.0 / 81.3 & \textbf{83.8} / \textbf{83.1} & 83.2 / 82.6\\
    \hline\hline
    \multicolumn{1}{|l|}{CR} & \textbf{79.7} / 78.0 & 78.0 / 77.3 & 78.6 / 77.0 & 79.1 / \textbf{78.2} \\
    \hline
    \multicolumn{1}{|l|}{MR} & \textbf{77.7} / \textbf{77.0} & 72.9 / 72.4 & 73.7 / 73.1 & 75.3 / 74.8\\
    \hline
    \multicolumn{1}{|l|}{MPQA} & \textbf{87.4} / 87.5 & 85.9 / 85.6 & 86.0 / 85.9 & \textbf{87.4} / \textbf{87.6}\\
    \hline
    \multicolumn{1}{|l|}{SUBJ} & \textbf{91.8} / \textbf{91.4} & 86.1 / 85.4 & 87.2 / 86.9 & 89.0 / 88.9\\
    \hline
    \multicolumn{1}{|l|}{SST Bin. Class.} & \textbf{80.4} / \textbf{81.4} & 78.1 / 77.5 & 77.2 / 76.7 & 78.1 / 78.8\\
    \hline
    \multicolumn{1}{|l|}{SST Fine-Grained Class.} & \textbf{45.1} / \textbf{44.4} & 38.3 / 40.5 & 40.5 / 39.3 & 41.9 / 41.4\\
    \hline
    \multicolumn{1}{|l|}{TREC} & \textbf{74.5} / \textbf{82.2} & 67.0 / 72.4 & 69.2 / 71.4 & 71.0 / 77.4\\
    \hline
    \multicolumn{1}{|l|}{MRPC} & \textbf{74.4} / 73.2 & 72.4 / 71.1 & 73.5 / 72.5 & 74.1 / \textbf{73.3}\\
    \hline
    \multicolumn{1}{|l|}{SICK-E} & 79.8 / 78.2 & 82.6 / 82.8 & \textbf{83.6} / 81.9 & 83.2 / \textbf{83.0}\\
    \hline\hline
	\multicolumn{1}{|l|}{Task (correlation)} & BoW & WRCE & Ours & Ours + BoW \\
	\hline
    \multicolumn{1}{|l|}{SICK-R} & 0.80 / 0.72 & 0.85 / 0.78 & \textbf{0.87} / \textbf{0.80} & 0.86 / \textbf{0.80}\\
    \hline
	\multicolumn{1}{|l|}{STS12} & 0.53 / 0.54 & 0.56 / 0.57 & 0.60 / 0.60 & \textbf{0.62} / \textbf{0.61}\\
    \hline
    \multicolumn{1}{|l|}{STS13} & 0.45 / 0.47 & 0.55 / 0.54 & 0.53 / 0.54 & \textbf{0.57} / \textbf{0.58}\\
    \hline
    \multicolumn{1}{|l|}{STS14} & 0.53 / 0.54 & 0.65 / 0.63 & 0.68 / \textbf{0.70} & \textbf{0.69} / 0.66\\
    \hline
    \multicolumn{1}{|l|}{STS15} & 0.56 / 0.59 & 0.68 / 0.69 & 0.70 / 0.70 & \textbf{0.71} / \textbf{0.72}\\
    \hline
    \multicolumn{1}{|l|}{STS16} & 0.52 / 0.57 & 0.69 / 0.70 & 0.70 / 0.72 & \textbf{0.71} / \textbf{0.73}\\
    \hline
\end{tabular}
\end{center}
\end{table*}

\subsection{Generalization}

We investigated which inputs influence correct predictions of the network
using the method of Integrated Gradients~\cite{Sundararajan2017Axiomatic}.
We have produced two heatmaps of input-output relationships for short (128 bytes)
and long (1024 bytes) paragraphs in our best model~(Figure~\ref{fig:heatmaps}).
In theory, a model performing identity should have a diagonal heatmap.
Our model finds relations within bytes of individual words, rarely crossing word and phrases
boundaries. In this sense, it fails to exploit the ordering of words.
However, the order is mostly preserved in the latent vector.
Early in the training the model learns to output only spaces,
which are the most common bytes in an average Wikipedia paragraph.
Later during training, it learns to correctly rewrite spaces,
while filling in the words with vowels, which are the most frequent non-space characters.
Interestingly, the compressing behavior seems to be language-specific and triggered
only by longer sequences. Figure~\ref{fig:samples} presents
input sentences and auto-encoded outputs of our best model,
trained on English Wikipedia, for English, French and random input sequences.

\section{Word-Level Sentence Encoder}
\label{sec:word-level-encoder}

Following the methods and work of~\cite{Conneau2017Supervised},
we apply our architecture to a practical task.
Namely, we train models consisting of the recursive convolutional word-level 
encoder and a simple three-layer fully-connected classifier on 
Stanford Natural Language Inference (SNLI) corpus~\cite{Bowman2015Large}. 
This dataset contains $570$k sentence pairs, each one described 
by one of three relation labels: entailment, contradiction, and neutral. 
Then we test encoders on various transfer tasks measuring 
semantic similarities between sentences.

The encoder of each model has a similar architecture
to the previously described byte-level encoder.
However, instead of bytes it takes words as its input sequence. 
Our best encoder has $N=8$ layers in each group.
The recursive group is applied $K$ times where $2^K$ is length of a padded
input sequence, so that the latent vector is of the size of a word vector. 
We use pre-trained GloVe vectors\footnote{\url{https://nlp.stanford.edu/projects/glove/}} 
and we do not fine-tune them.
We compared both fixed-length balanced, and variable length input paddings.
In fixed-length padding, up to first $64$ words are taken from each sentence.
We also compare ensemble of our best trained model and bag-of-words
as a sentence representation.
Let $v$ be the output vector of the encoder, and $u = \frac{1}{m} \sum_{i=1}^m e(w_i)$ be 
the average of word vectors of the sentence, where $m$ is the length of the sentence, 
$w_i$ is its $i$-th word, and $e(w)$ is the GloVe embedding of the word $w$.
Final embedding is the sum $x = v + u$.

Table~\ref{table:word-level} presents results for word-level
recursive convolutional encoder (WRCE), word-level model with fixed balanced padding (Ours),
and an ensemble of our model and an average embedding of the input sequence (Ours + BoW).
We compare them with a baseline model (BoW - average of GloVe vectors for words 
in a sentence) on SNLI and other classification tasks, SICK-Relatedness~\cite{Marelli2014SICK}, and STS\{$12$-$16$\} tasks. 
The \textit{SentEval}\footnote{\url{https://github.com/facebookresearch/SentEval}} 
tool was used for these experiments.

For certain tasks, especially those measuring textual similarity,
which are useful in retrieval-based response generation in dialogue systems,
presented models perform better than bag-of-words.
However, they are still not on par with LSTM-based methods~\cite{Conneau2017Supervised,Kiros2015Skip-Thought}
that generate more robust embeddings.
LSTM models are autoregressive and thus require slow sequential computations.
They are also larger, with the InferSent model~\cite{Conneau2017Supervised}
having over $30$ times more parameters than convolutional encoders presented in this section.
In addition, our architecture can share word embedding matrices with other
components of a conversational system, since word embeddings are ubiquitous in different modules of NLP systems.

In order to qualitatively assess how the results for those tasks transfer to the actual dialogue system,
we have compared some retrieved responses of a simple retrieval-based agent, which matches user
utterance with a single quote from Wikiquotes~\cite{Chorowski2018ATalker}.
We present a comparison of our word-level sentence encoder
with the bag-of-word method in response retrieval task (Figure~\ref{fig:knn}).
Human utterances from the training data
of NIPS 2017 Conversational Challenge\footnote{\url{http://convai.io/2017/data/}}
have been selected as input utterances.
We match them with the closest quote from Wikiquotes,
using a method similar to 
the one used in \textit{Poetwannabe} chatbot~\cite{Chorowski2018ATalker}.
All utterances have been filtered for foul speech (for details see~\cite{Chorowski2018ATalker}), tokenized using
Moses tokenizer\footnote{\url{https://github.com/moses-smt/mosesdecoder/blob/master/scripts/tokenizer/python-tokenizer/moses.py}},
and embedded as vectors.
For every user input utterance, we return the closest matching quote from Wikiquotes
with respect to the cosine similarity.

\section{Discussion and Future Work}
\label{sec:conclusion}

The problem of efficiently producing good and robust sentence embeddings
seems far from being solved.
In this work, which we consider a step forward in exploration of possible tools
and solutions,
we analyzed and proposed improvements to the BRCA model by Xiang Zhang and Yann LeCun~\shortcite{Zhang2018Byte-Level}.
With extensive usage of convolutions, our model is efficient in terms of computation and memory usage.
By analyzing BRCA we were able to diagnose problems with its training,
such as exploding gradients, and understand the difficulty in auto-encoding of long paragraphs,
especially in the initial stage of training.
Furthermore, we showed how to successfully apply batch normalization with recursive layers
and investigate input-output relations with Integrated Gradients method.

The recursive convolutional architecture benefits from the ease of training
and low number of parameters. Due to our realization that in the current byte-level
setting, input-output relations rarely cross word boundaries,
we demonstrate applicability of the
architecture in a word-level setting as a sentence embedder.
Furthermore, a good performance on semantic similarity tasks while using little resources
demonstrates it practical usefulness for dialog systems.

\section*{Acknowledgments}
The authors thank Paweł Rychlikowski and Michał Zapotoczny for fruitful discussions,
and Xiang Zhang for help towards a better understanding of the original model.
The research was partially supported by National Science Center (Poland) grant Sonata 8 2014/15/D/ST6/04402.

\begin{figure*}[h!]
\centering
{\footnotesize
\renewcommand{\arraystretch}{1.1}
\begin{tabular}{l l}
\hline
\rowcolor{mylightgray}
{\bf User}: & Hello, have you studied algebra? \\
{\bf BoW}: &  I never failed in mathematics. Before I was fifteen I had mastered differential and integral calculus. \\
{\bf Ours+BoW}: & I never failed in mathematics. Before I was fifteen I had mastered differential and integral calculus. \\
\hline
\rowcolor{mylightgray}
{\bf User}: & From what I understand it's a video game.\\
{\bf BoW}: &  Life is like a piano. What you get out of it depends on how you play it.\\
{\bf Ours+BoW}: & From now on, I control all aspects of the game.\\

\hline
\rowcolor{mylightgray}
{\bf User}: & God\\
{\bf BoW}: &  Knowledge of God is obedience to God.\\
{\bf Ours+BoW}: & God and all attributes of God are eternal.\\

\hline
\rowcolor{mylightgray}
{\bf User}: & Have you been to South Asia?\\
{\bf BoW}: &  We do not want chaos in South Africa.\\
{\bf Ours+BoW}: & There is a wide knowledge gap between us and the developed world in the West and in Asia.\\

\hline
\rowcolor{mylightgray}
{\bf User}: & How do you like Hackathon?\\
{\bf BoW}: &  Any thing you can do I can do better.\\
{\bf Ours+BoW}: & Discover the things that you do that make you proud as hell.\\

\hline
\rowcolor{mylightgray}
{\bf User}: & How do you think ?\\
{\bf BoW}: &  Any thing you can do I can do better.\\
{\bf Ours+BoW}: & How you think matters more than what you think.\\

\hline
\rowcolor{mylightgray}
{\bf User}: & I just didn't catch what was context message about.\\
{\bf BoW}: &  I read it because I was asked to explain what the truth is.\\
{\bf Ours+BoW}: & I spend so much time thinking about all the answers to the problem... that I forget what the problem actually was.\\

\hline
\rowcolor{mylightgray}
{\bf User}: & I'm an idiot\\
{\bf BoW}: &  I am an Agnostic because I am not afraid to think.\\
{\bf Ours+BoW}: & I wish I could say I was sorry.\\

\hline
\rowcolor{mylightgray}
{\bf User}: & It's classics!\\
{\bf BoW}: &  I love musical theatre and my dream is to do Once On This Island.\\
{\bf Ours+BoW}: & No work which is destined to become a classic can look like the classics which have preceded it.\\

\hline
\rowcolor{mylightgray}
{\bf User}: & So, start talking.\\
{\bf BoW}: &  Oh, ok, ok... Fair enough, yeah, rage it up. Rage all you want. Good things are coming. Good things.\\
{\bf Ours+BoW}: & Many people talk much, and then very many people talk very much more.\\

\hline
\rowcolor{mylightgray}
{\bf User}: & Technically correct\\
{\bf BoW}: &  Surely only correct understanding could lead to correct action.\\
{\bf Ours+BoW}: & Where an opinion is general, it is usually correct.\\

\hline
\rowcolor{mylightgray}
{\bf User}: & Thats why I play computer games alone.\\
{\bf BoW}: &  I have no time to play games.\\
{\bf Ours+BoW}: & The only legitimate use of a computer is to play games.\\

\hline
\rowcolor{mylightgray}
{\bf User}: & Well, can you dance?\\
{\bf BoW}: &  If I can mince , you can dance.\\
{\bf Ours+BoW}: & Ah, so you wish to dance.\\
\hline
\rowcolor{mylightgray}
{\bf User}: & What about ivy league?\\
{\bf BoW}: &  Ah wonder if anybody this side of the Atlantic has ever bought a baseball bat with playing baseball in mind.\\
{\bf Ours+BoW}: & This is so far out of my league.\\
\end{tabular}
}
\caption{Sample answers of retrieval-based agents which embed sentences as either BoWs, or BoWs combined with our method}
\label{fig:knn}

\end{figure*}

\vfill

\bibliographystyle{named}
\bibliography{refs}

\end{document}